\documentclass[journal]{IEEEtran}
\ifCLASSINFOpdf
\else
\fi

\usepackage{subfigure}
\usepackage{epsfig}
\usepackage{graphicx}
\usepackage{amsmath}
\usepackage{amssymb}
\usepackage{algorithm, algorithmic}
\graphicspath{{figure/}}   
\usepackage{multirow}

\usepackage{hyperref}
\usepackage{booktabs}
\usepackage{color}
\usepackage{bbding}
\begin{document}
%
\title{MultiScale Spectral-Spatial Convolutional Transformer for	Hyperspectral Image Classification}
%
%
%

\author{Zhiqiang~Gong,
			Xian~Zhou,
			Wen~Yao
\thanks{Manuscript received XX, 2023; revised XX, 2023. This work was supported by the Natural Science Foundation of China under Grant 62001502. 
}
\thanks{ Z. Gong and W. Yao are with the Defense Innovation Institute,
Chinese Academy of Military Sciences, Beijing 100000, China. e-mail: (gongzhiqiang13@nudt.edu.cn, wendy0782@126.com).}
\thanks{X. Zhou is with  the National Key Laboratory of Science and Technology on ATR, College of Electrical Science and Technology, National University of Defense Technology, Changsha 410073, China (e-mail: zhouxian@sjtu.edu.cn).}

}

%
%

\markboth{IEEE LATEX,~Vol X, 2023}%
{Shell \MakeLowercase{\textit{et al.}}: Bare Demo of IEEEtran.cls for IEEE Journals}
%



\maketitle

\begin{abstract}

Due to the powerful ability in capturing the global information, Transformer has become an alternative architecture of CNNs for hyperspectral image classification. However, general Transformer mainly considers the global spectral information while ignores the multiscale spatial information of the hyperspectral image.
In this paper, we propose a multiscale spectral-spatial convolutional Transformer (MultiFormer) for hyperspectral image classification. First, the developed method utilizes multiscale spatial patches as tokens to formulate the spatial Transformer and generates multiscale spatial representation of each band in each pixel. Second, the spatial representation of all the bands in a given pixel are utilized as tokens to formulate the spectral Transformer and generate the multiscale spectral-spatial representation of each pixel. 
Besides, a modified spectral-spatial CAF module is constructed in the MultiFormer to fuse cross-layer spectral and spatial information.
Therefore, the proposed MultiFormer can capture the multiscale spectral-spatial information and provide better performance than most of other architectures for hyperspectral image classification.
Experiments are conducted over commonly used real-world datasets and the comparison results show the superiority of the proposed method.
\end{abstract}


%
\IEEEpeerreviewmaketitle

\section{Introduction}

Hyperspectral imaging, as a powerful remote sensing technique, can caputure information across a broader range of the electromagnetic spectrum, including infrared and ultraviolet spectra \cite{b17}. This enables the obtained hyperspectral images to delve deeper into the spectral characteristics of surface objects and make significant breakthroughs in various fields such as agriculture \cite{b18}, environmental monitoring \cite{b19}, and military applications \cite{b20}. Hyperspectral image classification, which tries to categorizing each pixel into predefined classes based on their spectral signatures, plays a crucial role in various applications such as land cover mapping \cite{b21}, environmental monitoring \cite{b22}, mineral exploration \cite{b23}, and ruban planning \cite{b24}. However, the great similarity and overlapping between the spectral signature of different objects make it difficult to discriminate different objects. 
Faced these circumstances, exploring effective spectral-spatial methods to incorporate the spatial information for hyperspectral image classification have become increasingly important.

Nowadays, advanced deep leanring methods have demonstrated their ability to discriminate subtle spectral difference of different objects in hyperspectral image. Among these methods, convolutional neural networks (CNNs) are the most representative ones \cite{b25}. Generally, these methods possess multiple layers with large amounts of parameters and learn these parameters adaptively. Besides, due to the specific convolution operation, the CNNs have strong potential to capture the local correlation. As a result, the CNNs can extract subtle spectral-spatial features to discriminate different objects. Much efforts have been conducted on better CNN backbones and their variants to improve the performance of CNNs for hyperspectral image classification \cite{gong1,gong2}. However, limited by the architectures of CNNs, CNNs present slightly weak ability to capture long-range correlation which is quite important for hyperspectral image processing since the images usually have hundreds of bands. 

As an alternative of CNNs, the transformer has become another paradigm of deep learning architectures. It generally consists of self-attention mechanism, positional encoding, residual connections, and others, and enables the Transformer to capture long-range dependencies and relationships \cite{b10}. 
Due to the good performance for extracting global features, Transformer has been widely applied in natural language processing. Some variants have also been developed to expand the application domains and further enhance its ability to extract multi-scale information, hierarchical information, and other capabilities, such as Vision Transformer \cite{b1}, Transformer in Transformer \cite{b2}, swin transformer \cite{b15}, DeiT \cite{b16}. It has also been applied in the literature of hyperspectral image classification, such as SpectralFormer\cite{b3}, SSFTT \cite{b4}. SpectralFormer mainly constructs the tokens with spectral embeddings and cannot fully utilize the spatial information. The SSFTT ignores the mutliscale spatial information of hyperspectral image. Faced these problems in prior transformers for current task, this work will propose a novel Transformer which can fully utilize both the spectral and spatial information.

As for hyperspectral image, it contains plentiful spectral and spatial information. Most of prior works mainly consider the use of spectral information of the neighbors to utilize the spectral-spatial information while underutilizing the inherent spatial information. Multi-scale is validated as one of important characteristics of hyperspectral images and prior works have already demonstrated their effectiveness to use the multi-scale spatial information for hyperspectral image classification \cite{gong3, b26, b27, b28, b29}. 
Wang et al. \cite{b27} developed a unified multiscale learning (UML) framework, which consists of a multiscale spatial-channel attention mechanism and a multiscale shuffle block to improve the problem of land-cover map distortion.
Wan et al. \cite{b28} proposed a MDGCN, which can establish multiple input graphs with
different neighborhood scales to extensively exploit the diversified
spectral-spatial correlations at multiple scales, to comprehensively deploy the multi-scale information inherited by
hyperspectral images.
For example, Fang et al. \cite{b29} proposed a multiscale adaptive sparse representation (MASR) model to effectively exploit spatial information at multiple scales via an adaptive sparse strategy.
To take advantage of such multi-scale spatial information,
this work will develop a novel multiscale spatial convolutional transformer block for better classification performance.

Considering the merits of spectral-spatial transformer and multiscale spatial information, this work will develop a novel multiscale spectral-spatial convolutiuonal Transformer for hyeprspectral image classification. First, multiscale spatial information of each band is extracted as tokens through convolution operation to construct the multiscale spatial convolutional Transformer, aiming at extracting the spatial information of the hyperspectral image. Then, the spectral transformer is constructed by using obtained multiscale information of each band as tokens. Besides, local recycle module is introduced to fuse the cross-layer information to further improve the representational ability of the model.
To be concluded, this paper makes the following contributions.

\begin{itemize}
\item This work proposes a novel MultiscaleFormer through embedding multiscale spatial convolutional transformer in spectral transformer which can fully utilize the spectral information and the multiscale spatial information.
\item This work develops the multiscale spatial convolutional transformer, which takes the multiscale spatial information of each band as tokens, to fully utilize the multiscale spatial information of hyperspectral image.
\item Motivated by the cross-layer adaptive fusion module \cite{b3}, this work constructs the spectral-spatial cross-layer fusion module to fuse both the cross-layer information of multiscale spatial convolutional transformer and spectral tranformer in the proposed MultiscaleFormer to further improve the representational ability.
\end{itemize}

The remainder of this paper is arranged as follows. Section II details the proposed multiscale spatial-spectral convolutional transformer for hyperspectral image classification. In Section III, experimental results over commonly used real-world datasets demonstrate the effectiveness of proposed method. Finally, Section IV summarizes this paper with some future directions.  

\section{Proposed Method}
In this section, we will describe the proposed MultiscaleFormer in detail, including the overall framework, the multiscale spatial convolutioanl transformer, the spectral transfromer, and the spectral-spatial CAF module.

\subsection{Overall MultiscaleFormer Framework}

\begin{figure*}[ht]
\centering
\includegraphics[width=0.99\linewidth]{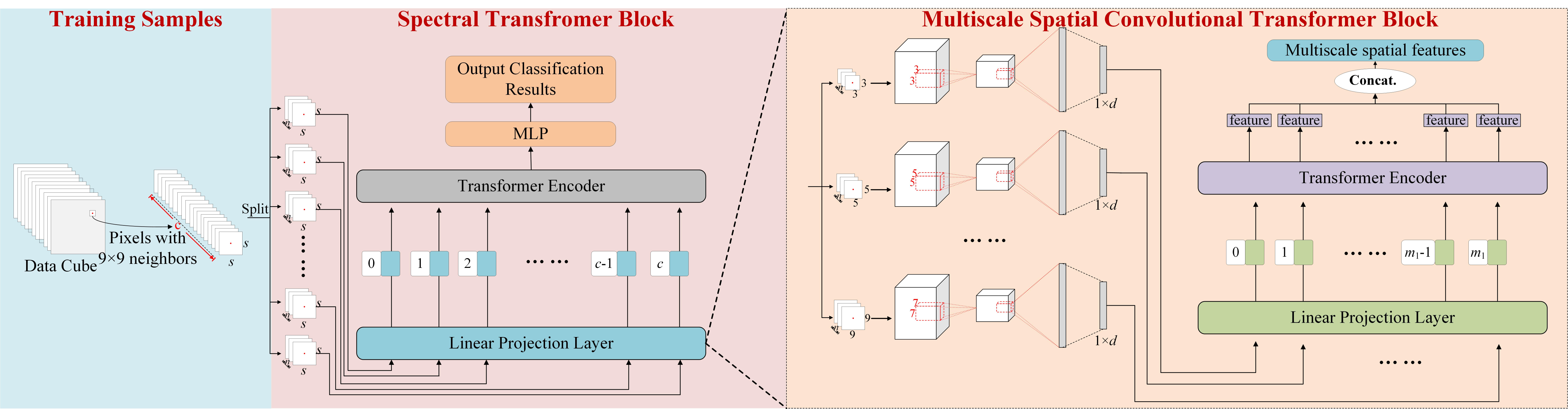}
  \caption{The overall framework of the proposed multiscale spectral-spatial convolutional transformer for hyperspectral image classification.}
\label{fig:framework}
\end{figure*}

To fully utilize the multiscale spatial information and spectral information, this paper develops the MultiFormer which embeds multiscale spatial convolutional transformer in spectral transformer.
The overall framework of the proposed MultiscaleFormer is showed in Fig. \ref{fig:framework}. Generally, the MultiFormer is formulated based on basic Transformer, and consists of three parts: multiscale spatial convolutional transformer as inner transformer block, the spectral transformer as outer transformer block, and the spectral-spatial cross-layer adaptive fusion module. The inner transformer block, e.g. the proposed multiscale spatial convolutional transformer, takes advantage of the multiscale spatial information to construct tokens and formulate the specific tranformer to extract multiscale spatial information of hyperspectral image. The outer transformer block, e.g. the spectral transformer, takes the extracted spatial features of each band as tokens to formulate the spectral transformer to extract discriminative features. 
By stacking the inner and outer blocks for $L$ times, we build the MultiscaleFormer. 
Finally, the spectral-spatial cross-layer adaptive fusion module is constructed and applied in both the Multiscale Spatial Convolutional Transformer and the Spectral Transformer to fuse the cross-layer spatial and spectral information, respectively. 

\subsection{Basic Transformer}
We first briefly describe the basic components in basic transformer \cite{b10}, including MSA (Multi-head Self-Attention), MLP (Multi-Layer Perceptron) and LN (Layer Normalization).

\subsubsection{MSA}

In the self-attention module, the inputs $X\in \mathbb{R}^{n\times d}$ are linearly transformed to three parts, i.e., queries $Q\in \mathbb{R}^{n\times d_k}$, keys $K\in \mathbb{R}^{n\times d_k}$ and values $V\in \mathbb{R}^{n\times d_v}$ where $n$ is the sequence length, $d$, $d_k$, $d_v$ are the dimensions of inputs, queries (keys) and values, respectively. The scaled dot-product attention is applied on $Q, K, V$:
\begin{equation}
Attention(Q, K, V)=softmax(\frac{QK^T}{\sqrt{d_k}})V.
\end{equation}
Finally, a linear layer is used to produce the output. Multi-head self-attention splits the queries, keys and values to $h$ parts and perform the attention function in parallel, and then the output values of each head are concatenated and linearly projected to form the final output. The Multi-head self-attention can be represented as
\begin{equation}
\begin{aligned}
MSA(Q, K, V)=Concat(head_1, \cdots, head_h)W^O,  \\
\mathrm{ s.t. } \quad head_i = Attention(QW_i^Q, KW_i^K, VW_i^V),
\end{aligned}
\end{equation}
where $head_1, \cdots, head_h$ denotes the $h$ heads, and $W_i^Q, W_i^K, W_i^V$, and $W^O$ are the projection weights in multi-head self-attention.

\subsubsection{MLP}
The MLP is applied between self-attention layers for ffeature transformation and non-linearity:
\begin{equation}
MLP(X)=FC(\sigma(FC(X))), FC(X)=XW+b,
\end{equation}
where $W$ and $b$ are the weight and bias term of fully-connected layer respectively, and $\sigma(\cdot)$ is the activation function such as GELU\cite{b11}.

\subsubsection{LN}
Layer normalization \cite{b12} is a key part in transformer for stable training and faster convergence. LN is applied over each samples $x\in \mathbb{R}^d$ as follows:
\begin{equation}
LN(x)=\frac{x-\mu}{\delta}\circ\gamma+\beta,
\end{equation}
where $\mu\in \mathbb{R}$, $\delta\in \mathbb{R}$ are the mean and standard deviation of the feature respectively, $\circ$ is the element-wise dot, and $\gamma\in \mathbb{R}^d$, $\beta\in \mathbb{R}^d$ are learnable affine transform parameters.

\subsection{Multiscale Spatial Convolutional Transformer Block}
\subsubsection{Multiscale Spatial Embeddings}
Multi-scale feature extraction methods in the context of hyperspectral images have been proven to be effective in improving classification performance \cite{gong3, b13}. Such methods allow the model to process hyperspectral images at multiple scales, accommodating variations in spatial detail and spectral characteristics, and can effectively handle complex scenes with objects or materials that may have diverse scales and contextual features, making it well-suited for hyperspectral image classification tasks.
For this purpose, we propose to learn multiscale spatial embeddings, i.e., Multiscale Spatial Convolutional Transformer block, which combines the strengths of convolutional neural networks (CNNs) for spatial feature extraction with the self-attention mechanisms of transformer blocks to capture long-range dependencies and relationships within the data.

\begin{figure}[t]
\centering
\includegraphics[width=0.99\linewidth]{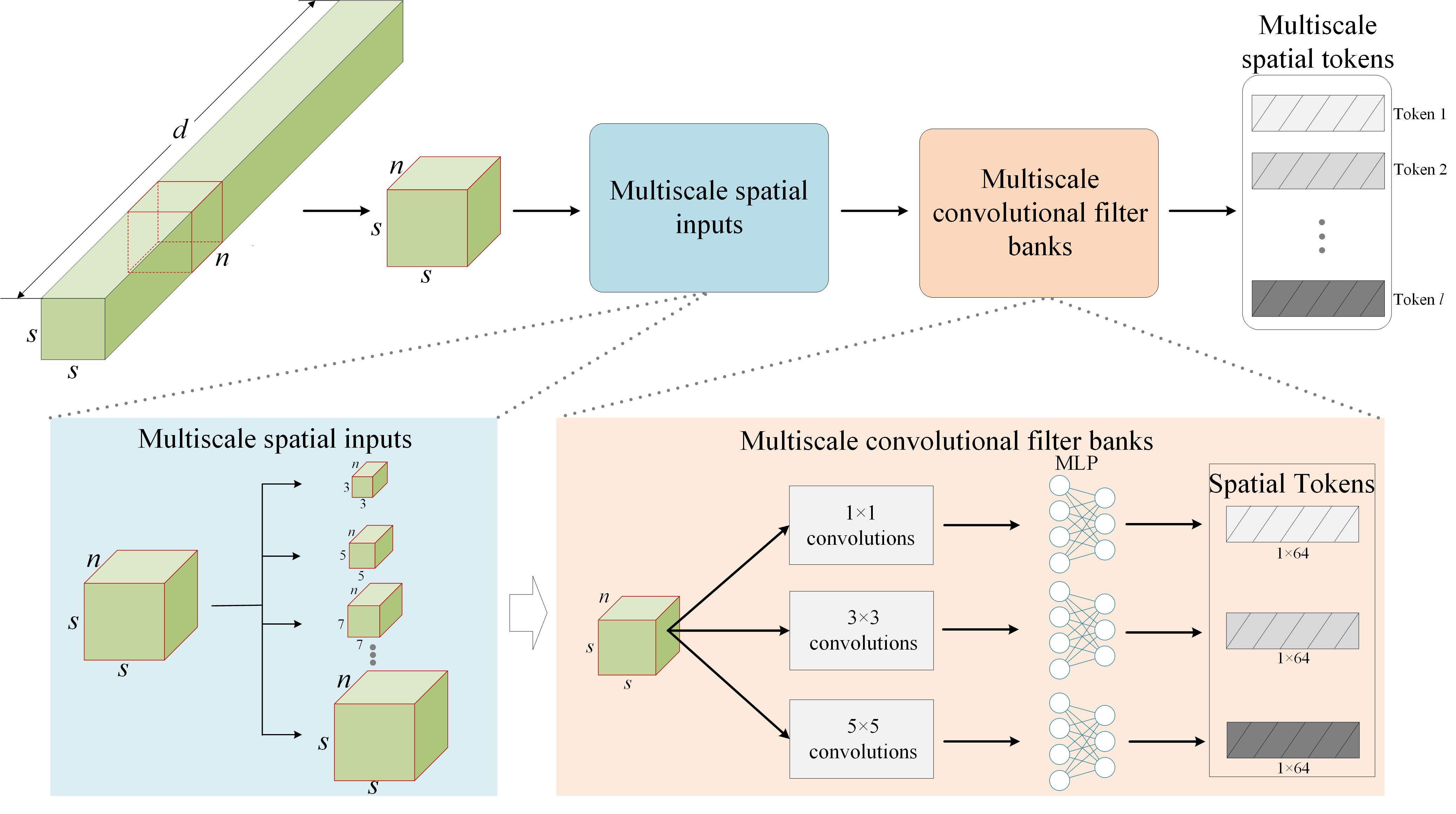}
  \caption{An illustration of Multiscale Spatial Embeddings in Multiscale Spatial Convolutional Transformer.}
\label{fig:embedding}
\end{figure}

The key to utilize the multiscale spatial information and formulate the specific transformer block is to construct the multiscale spatial tokens.
This work combines the multi-scale spatial sizes and multi-scale convolutional filter banks given a specific spatial size to formulate the multiscale spatial tokens. Detailed illustration of multiscale spatial embeddings can be seen in Fig. \ref{fig:embedding}.

The multiscale spatial neighbors of a pixel refer to the surrounding pixels within different spatial neighborhoods or scales.
By considering spatial information at multiple scales, classifiers can account for various spatial structures, contextual information, and relationships among neighboring pixels.
As shown in Fig. \ref{fig:embedding}, this work uses $3\times 3, 5\times 5, 7\times 7, \cdots, s\times s$ spatial neighbors to construct the multiscale spatial inputs.

The multi-scale
filter bank conceptually similar to the inception module in [12]
is used to optimally exploit diverse local structures of the input
image.
They consist of a collection of convolutional filters, each designed to capture specific patterns or structures within the data, and these filters operate at different scales or receptive field sizes. By employing multiscale convolutional filter banks, we can efficiently process and analyze complex data, and uncover a hierarchy of information.  by providing a versatile means, the multi-scale filter filter bank can extract relevant features and representations from data across a wide range of scales and frequencies.

In the multiscale spatial convolutinal transformer block, the multiscale spatial inputs are first extracted, and then the multi-scale filter banks are followed to construct the multiscale spatial embeddings.
For inputs with different spatial neighbors, multi-scale filter bank locally convolves
the input image with several convolutional filters with different
sizes (e.g., $3\times 3$, $5\times 5$, and $7\times 7$) and strides.

Denote ${\bf x}\in \mathbb{R}^{s\times s\times c}$ as a training sample from the hyperspectral image and $y$ is the corresponding label of ${\bf x}$, where $s$ represents the spatial size used for classification (i.e. $9\times 9$ neighbors) and $c$ stands for the number of spectral bands.

For a specific band $j\in \{1,2,\cdots,c\}$, first we obtain the multiscale spatial embedding ${\bf z}^{j} \in \mathbb{R}^{m_1\times d_1}$ given ${\bf x}$ (as Fig. \ref{fig:embedding} shows), where $m_1$ denotes the number of multiscale spatial representations and $d_1$ is the dimension of multiscale spatial embedding.
Then, the multiscale spatial embedding ${\bf z}^{j}$ can represented as a sequence of spatial embeddings: 

\begin{equation}
{\bf z}^j\rightarrow [{\bf z}^{j,1}, {\bf z}^{j,2}, \cdots, {\bf z}^{j, m_1}]
\end{equation}

\subsubsection{Inner Transformer Block}

Then, we utilize the formulate the transformer block, s.t., multiscale spatial convolutional transformer block, to explore the relation between these multiscale spatial embeddings.
\begin{equation}
\hat{{\bf z}}_l^j={\bf z}_{l-1}^j+MSA(LN({\bf z}_{l-1}^j))
\end{equation}
\begin{equation}
{{\bf z}}_l^j=\hat{{\bf z}}_{l}^j+MLP(LN(\hat{{\bf z}}_{l}^j))
\end{equation}
where $l\in\{1,2,\cdots, L\}$ is the $l$-th block and $L$ denotes the total number of stacked blocks.

\subsection{Spectral Transformer Block}

\subsubsection{Spectral Embeddings}
For MultiscaleFormer, the spectral Transformer block takes the multiscale spatial features extracted from the multiscale spatial convolutional transformer block as spectral tokens.

First, create the spectral embedding memories to store the sequence of spectral-level representations:
\begin{equation}
{{\bf p}}=[{{\bf p}}^0, {{\bf p}}^1, {{\bf p}}^2, \cdots, {{\bf p}}^c]\in \mathbb{R}^{(c+1)\times d_2}
\end{equation}
where ${{\bf p}}^0$ represents the class token as \cite{b1}, $d_2$ is the dimension of spectral embedding, and ${\bf p}$ is initialized as zero.

In each layer, the sequence of multiscale spatial embeddings are transformed into the domain of spectral embedding by linear projection and added into the multiscale spatial embedding:
\begin{equation}
{{\bf p}}_{l}^j={{\bf p}}_{l}^j+FC(Vec({{\bf z}}_l^j)),
\end{equation}
where ${{\bf p}}_{l}^j\in \mathbb{R}^d$ and FC, which is the fully-connected layer, makes the dimension match for addition. $Vec(\cdot)$ is the vectorization operation. 

\subsubsection{Outer Transformer Block}
We use the standard transformer block for transforming the spectral embeddings:
\begin{equation}
\hat{{{\bf p}}}_l={{\bf p}}_{l-1}+MSA(LN({{\bf p}}_{l-1})),
\end{equation}
\begin{equation}
{{\bf p}}_l=\hat{{{\bf p}}}_l+MLP(LN(\hat{{{\bf p}}}_l)).
\end{equation}
This outer spectral transformer block is used for modeling relationships among spectral embeddings.

\subsection{Spectral-Spatial Cross-Layer Adaptive Fusion Module}

\begin{figure*}[ht]
\centering
\includegraphics[width=0.9\linewidth]{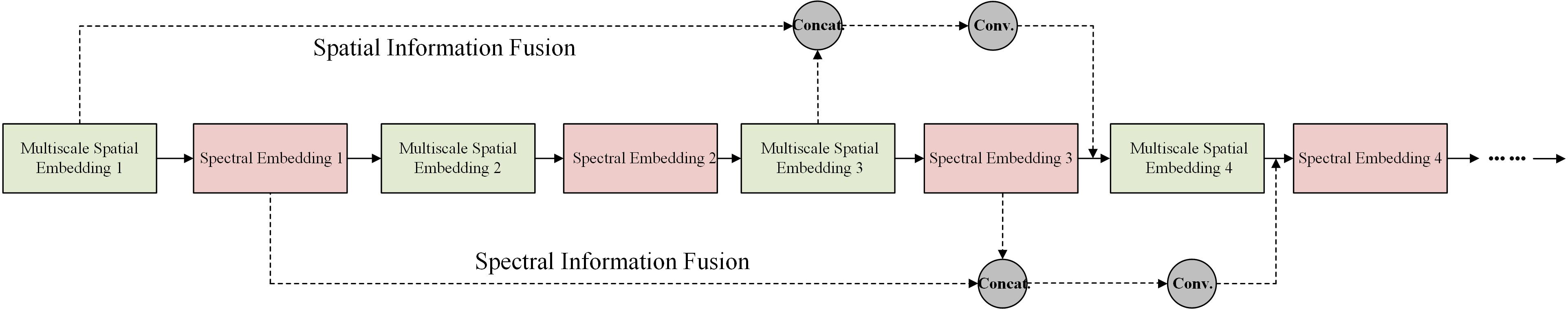}
  \caption{The Spectral-Spatial Cross-Layer Adaptive Fusion Module.}
\label{fig:caf}
\end{figure*}

In deep networks, particularly when stacking multiple layers, there is a risk of information degradation as the input data passes through successive non-linear transformations. Skip connections address this issue by providing a direct pathway for information to flow, ensuring that valuable features and information from the input data are preserved \cite{b14}.
To this end, motivated by \cite{b3}, this work designs the spectral-spatial cross-layer adaptive fusion module.
Fig. \ref{fig:caf} shows the detailed illustration of the module.

As shows in the figure, the module consists of the spatial information fusion part and spectral information fusion part over the multiscale spatial convolutional transformer block and the spectral transformer block, respectively.
As for multiscale spatial information fusion, denote ${{\bf z}}_{l-2}^j\in \mathbb{R}^{m_1\times d_1}$ and ${{\bf z}}_{l}^j\in \mathbb{R}^{m_1\times d_1}$ be the outputs (or representations) in the $(l-2)$-th and $(l)$-th blocks, respectively, the SCAF can be then expressed by
\begin{equation}
\hat{{\bf z}}_{l}^j \leftarrow \hat{w}\binom{{{\bf z}}_{l-2}^j}{{{\bf z}}_{l}^j}
\end{equation}
where $j$ denotes the specific spectral band, $\hat{{\bf z}}_{l}^j$ denotes the fused representations in the $(l)$-th layer with the SCAF, and $\hat{w}\in \mathbb{R}^{1\times 2}$ is the learnable network parameter for adaptive fusion.

Similarly, As for the spectral information fusion, denote ${{\bf p}}_{l-2}\in \mathbb{R}^{(c+1)\times d_2}$ and ${{\bf p}}_{l}\in \mathbb{R}^{(c+1)\times d_2}$ be the outputs (or representations) in the $(l-2)$-th and $(l)$-th layers, respectively, the SCAF can be then expressed by
\begin{equation}
\hat{{\bf p}}_{l} \leftarrow \hat{v}\binom{{{\bf p}}_{l}}{{{\bf p}}_{l-2}}
\end{equation}
where $\hat{{\bf p}}_{l}$ denotes the fused representations in the $(l)$-th layer with the SCAF, and $\hat{v}\in \mathbb{R}^{1\times 2}$ is the learnable network parameter for adaptive fusion.

\section{Experimental Results}

\subsection{Experimental Datasets and Setups}
\subsubsection{Experimental Datasets}
To  evaluate the performance of the proposed method, this work further conducts experiments over two common used ureal-world hyperspectral images, namely the Indian Pines data \cite{data} and the Houston 2013 data \cite{houston2013}.


{\bf Indian Pines (IP) data} was gathered by the Airborne Visible/Infrared Imaging Spectrometer (AVIRIS) sensor over the Indian PInes test set in Northwestern Indiana. It consists of $145\times 145$ pixels with spectral bands  ranging from 0.4 to 2.5 $\mu$m. 24 bands covering the region of water absorption are removed and the remaining 200 spectral bands are used. 16 land cover classes with a total of 10249 labeled samples are selected for experiments. Table \ref{table:indian} shows the training and testing samples in the experiments.

{\bf Houston 2013 (HS) data } was collected by the National Center for Airborne Laser Mapping (NCALM) over the University of Houston campus and the neighboring urban area  throuth ITRES CASI 1500 sensor at the spatial resolution of 2.5m. The cube consists of $349\times 1905$ pixels with 144 spectral bands ranging from 380 nm to 1050 nm. 15 land cover classes with a total of 15029 labeled samples are selected for experiments. The training and testing samples are listed detailedly in Table \ref{table:houston2013}.

\begin{table}[t]
\begin{center}
\caption{Number of training and testing samples in Indian Pines data.}
\label{table:indian}
\begin{tabular}{ c | c c c c }
\toprule[1pt]
{Class}     &  {Class Name} & Color &  {Training}&  {Testing}  \\
\hline\hline
C1   &  Corn-notill                    &\begin{minipage}[b]{0.08\columnwidth}
		\raisebox{-.45\height}{\includegraphics[width=\linewidth]{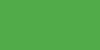}}
	\end{minipage} & 50  & 1384   \\
C2   &  Corn-mintill                   &\begin{minipage}[b]{0.08\columnwidth}
		\raisebox{-.45\height}{\includegraphics[width=\linewidth]{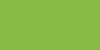}}
	\end{minipage} & 50  & 784  \\
C3   &  Corn                           &\begin{minipage}[b]{0.08\columnwidth}
		\raisebox{-.45\height}{\includegraphics[width=\linewidth]{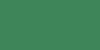}}
	\end{minipage} & 50  & 184   \\
C4   &  Grass-pasture                  &\begin{minipage}[b]{0.08\columnwidth}
		\raisebox{-.45\height}{\includegraphics[width=\linewidth]{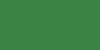}}
	\end{minipage} & 50  & 447   \\
C5   &  Grass-trees                    &\begin{minipage}[b]{0.08\columnwidth}
		\raisebox{-.45\height}{\includegraphics[width=\linewidth]{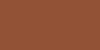}}
	\end{minipage} & 50  & 697  \\
C6   &  Hay-windrowed                  &\begin{minipage}[b]{0.08\columnwidth}
		\raisebox{-.45\height}{\includegraphics[width=\linewidth]{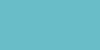}}
	\end{minipage} & 50  & 439   \\
C7   &  Soybean-notill                 &\begin{minipage}[b]{0.08\columnwidth}
		\raisebox{-.45\height}{\includegraphics[width=\linewidth]{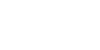}}
	\end{minipage} & 50  & 918   \\
C8   &  Soybean-mintill                &\begin{minipage}[b]{0.08\columnwidth}
		\raisebox{-.45\height}{\includegraphics[width=\linewidth]{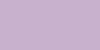}}
	\end{minipage} & 50  & 2418   \\
C9   &  Soybean-clean                  &\begin{minipage}[b]{0.08\columnwidth}
		\raisebox{-.45\height}{\includegraphics[width=\linewidth]{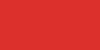}}
	\end{minipage} & 50  & 564  \\
C10   & Wheat                          &\begin{minipage}[b]{0.08\columnwidth}
		\raisebox{-.45\height}{\includegraphics[width=\linewidth]{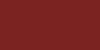}}
	\end{minipage} & 50  & 162  \\
C11   & Woods                          &\begin{minipage}[b]{0.08\columnwidth}
		\raisebox{-.45\height}{\includegraphics[width=\linewidth]{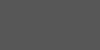}}
	\end{minipage} & 50  & 1244  \\
C12   & Buildings-Grass-Trees-Drives   &\begin{minipage}[b]{0.08\columnwidth}
		\raisebox{-.45\height}{\includegraphics[width=\linewidth]{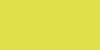}}
	\end{minipage} & 50  & 330  \\
C13   & Stone-Steel-Towers             &\begin{minipage}[b]{0.08\columnwidth}
		\raisebox{-.45\height}{\includegraphics[width=\linewidth]{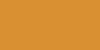}}
	\end{minipage} & 50  & 45  \\
C14   & Alfalfa                        &\begin{minipage}[b]{0.08\columnwidth}
		\raisebox{-.45\height}{\includegraphics[width=\linewidth]{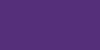}}
	\end{minipage} & 15  & 39  \\
C15   & Grass-pasture-mowed            &\begin{minipage}[b]{0.08\columnwidth}
		\raisebox{-.45\height}{\includegraphics[width=\linewidth]{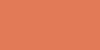}}
	\end{minipage} & 15  & 11  \\
C16   & Oats                           &\begin{minipage}[b]{0.08\columnwidth}
		\raisebox{-.45\height}{\includegraphics[width=\linewidth]{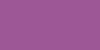}}
	\end{minipage} & 15  & 5  \\
 \hline\hline
Total     &                            & & 695 & 9671  \\
\bottomrule[1pt]
\end{tabular}
\end{center}
\end{table}

\begin{table}[t]
\begin{center}
\caption{Number of training and testing samples in Houston2013 data.}
\label{table:houston2013}
\begin{tabular}{ c | c c c c }
\toprule[1pt]
{Class}     &  {Class Name} & Color&  {Training} &  {Testing} \\
\hline\hline
C1   &  Grass-healthy   &  \begin{minipage}[b]{0.08\columnwidth}
		\raisebox{-.45\height}{\includegraphics[width=\linewidth]{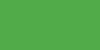}}
	\end{minipage}& 198  & 1053   \\
C2   &  Grass-stressed  &  \begin{minipage}[b]{0.08\columnwidth}
		\raisebox{-.45\height}{\includegraphics[width=\linewidth]{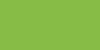}}
	\end{minipage}& 190  & 1064  \\
C3   &  Grass-synthetic &  \begin{minipage}[b]{0.08\columnwidth}
		\raisebox{-.45\height}{\includegraphics[width=\linewidth]{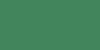}}
	\end{minipage}& 192  & 505   \\
C4   &  Tree            & \begin{minipage}[b]{0.08\columnwidth}
		\raisebox{-.45\height}{\includegraphics[width=\linewidth]{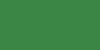}}
	\end{minipage} & 188  & 1056   \\
C5   &  Soil            &  \begin{minipage}[b]{0.08\columnwidth}
		\raisebox{-.45\height}{\includegraphics[width=\linewidth]{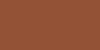}}
	\end{minipage}& 186  & 1056  \\
C6   &  Water           &  \begin{minipage}[b]{0.08\columnwidth}
		\raisebox{-.45\height}{\includegraphics[width=\linewidth]{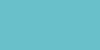}}
	\end{minipage}& 182  & 143   \\
C7   &  Residential     &  \begin{minipage}[b]{0.08\columnwidth}
		\raisebox{-.45\height}{\includegraphics[width=\linewidth]{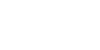}}
	\end{minipage}& 196  & 1072   \\
C8   &  Commercial      &  \begin{minipage}[b]{0.08\columnwidth}
		\raisebox{-.45\height}{\includegraphics[width=\linewidth]{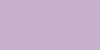}}
	\end{minipage}& 191  & 1053   \\
C9   &  Road            &  \begin{minipage}[b]{0.08\columnwidth}
		\raisebox{-.45\height}{\includegraphics[width=\linewidth]{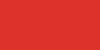}}
	\end{minipage}& 193  & 1059  \\
C10   & Highway         &  \begin{minipage}[b]{0.08\columnwidth}
		\raisebox{-.45\height}{\includegraphics[width=\linewidth]{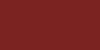}}
	\end{minipage}& 191  & 1036  \\
C11   & Railway         &  \begin{minipage}[b]{0.08\columnwidth}
		\raisebox{-.45\height}{\includegraphics[width=\linewidth]{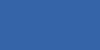}}
	\end{minipage}& 181  & 1054  \\
C12   & Parking-lot1    &  \begin{minipage}[b]{0.08\columnwidth}
		\raisebox{-.45\height}{\includegraphics[width=\linewidth]{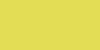}}
	\end{minipage}& 192  & 1041  \\
C13   & Parking-lot2    &  \begin{minipage}[b]{0.08\columnwidth}
		\raisebox{-.45\height}{\includegraphics[width=\linewidth]{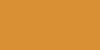}}
	\end{minipage}& 184  & 285  \\
C14   & Tennis-court    & \begin{minipage}[b]{0.08\columnwidth}
		\raisebox{-.45\height}{\includegraphics[width=\linewidth]{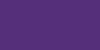}}
	\end{minipage} & 181  & 247  \\
C15   & Running-track   &  \begin{minipage}[b]{0.08\columnwidth}
		\raisebox{-.45\height}{\includegraphics[width=\linewidth]{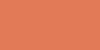}}
	\end{minipage}& 187  & 473 \\

 \hline\hline
Total     &             & & 2832 & 12197  \\
\bottomrule[1pt]
\end{tabular}
\end{center}
\end{table}

%

\subsubsection{Experimental Setups}

All the experiments in the paper are implemented under Pytorch 1.9.1. Very common machine with a Intel@ Xeon(R) Gold 6226R CPU, 128GB RAM and Quadro RTX 6000 24GB GPU is used to evaluate the classification performance. $9\times 9$ spatial neighbors are used as input to incorporate the spatial information. 

{\bf Evaluation Metrics}
The classification performance is evaluated under three commonly used metrics, namely Overall Accuracy (OA), Average Accuracy (AA), and Kappa coefficient ($\kappa$). Besides, the classifcation accuracy per class is provided to make a thorough comparison.

{\bf Baseline Methods}
This work selects SVM, 3-D CNN \cite{b8},
PResNet \cite{b7},
HybridSN \cite{b6},
RNN \cite{b9},
miniGCN \cite{b5},
ViT \cite{b1}, SpectralFormer \cite{b3}, SSFTTNet \cite{b4} as baseline methods to validate the effectiveness of the proposed method. Among these methods, SVM is chosen as representive of traditional handcrafted methods. 3-D CNN, PResNet, and HybridSN are chosen as representatives of advanced CNNs. ViT, SpectralFormer, and SSFTTNet are selected as the state-of-the-art Transformers for current tasks.

\subsection{General Performance}

At first, we present a brief overview of the merits of the proposed MultiFormer for hyperspectral image classification.
In this set of experiments,
the number of patch dimension, spectral neighbors, model layer is set to 64, 5, 5, respectively.
The proposed method contains the multiscale spatial convolutional transformer, the spectral transformer, and the spectral-spatial cross-layer adaptive fusion module. This part will test the ablation study of the proposed method over both the datasets. Table \ref{table:ablation} lists the comparison results where MS denotes the multiscale spatial convolutional transformer and SCAF denotes the spectral-spatial cross-layer adaptive fusion module.

From the results in the table, we can find that both the  multiscale spatial convolutional transformer and the spectral-spatial cross-layer adaptive fusion module can boost the classification performance. As for Indian Pines data, the proposed method with the multiscale spatial convolutional transformer can obtain the accuracy of 82.87\% OA which is better than 65.16\% OA while the proposed method with both the multiscale spatial convolutional transformer and the spectral-spatial cross-layer adaptive fusion module can further improve the performance by 0.97\%. While for Houston2013 data, the proposed method can also obtain an improvement by 0.91\%, 6.16\% than that without  spectral-spatial cross-layer adaptive fusion module and that without both the module.

\begin{table}[t]
\begin{center}
\caption{Classification accuracies (OA, AA, and $\kappa$) of the ablation study of the proposed method over different datasets.}
\label{table:ablation}
\begin{tabular}{ c | c c || c c c c }
\toprule[1pt]
{ Data}     &    {MS}&    { SCAF}&  { OA(\%)}&  {AA(\%)} &  { $\kappa$(\%)}  \\
\hline\hline
  \multirow{3}{*}{IP}    & \XSolidBrush &  \XSolidBrush & 65.16 &70.21 & 60.26 \\
       & \Checkmark  &\XSolidBrush &82.87 &89.46  &80.50  \\
       & \Checkmark  &\Checkmark &{\bf 83.84} &{\bf 90.58}  & {\bf 81.61}   \\
\hline
 \multirow{3}{*}{HS}      &  \XSolidBrush   &\XSolidBrush &82.22 &83.43 & 80.68  \\
      & \Checkmark  &\XSolidBrush & 83.13 & 83.45 & 81.69  \\
			& \Checkmark  &\Checkmark & {\bf 88.38}&{\bf 88.21} & {\bf 87.38}  \\
\bottomrule[1pt]
\end{tabular}
\end{center}
\end{table}

\subsection{Performance with Different Number of  Model Blocks $L$}

Apart from the learnable parameters in networks and hyper-parameters required in the trainig process, the number of model layers of the proposed MultiFormer plays a vital role in the final classification performance.
Therefore, it is indispensable to explore the proper paramaeter range. Similarly, we investigate the performance with different number of model layers over both the datasets. Table \ref{table:layers} lists the changing trend of the classification accuracies with the gradual increase of the number of model layers in terms of OA, AA, and $\kappa$.

A common conclusion is that the proposed MultiFormer with more layers have higher representational ability and can provide better classification performance. However, due to the limited training samples and training strategies, the performance with excessive large number of layers is slightly decreased.
Over Indian Pines data, the proposed method can obtain an accuracy of 86.25\% OA when the number of layers is set to 7 which performs the best. Besides, the accuracy decreased to 84.97\% and 85.93\% when the number is set to 8 and 9, respectively. As for Houston2013 data, the proposed method performs the best (88.38\% OA) when the number is set to 5.

\begin{table*}[t]
\begin{center}
\caption{Classification accuracies (OA, AA, and $\kappa$) of the proposed method with different number of layers.}
\label{table:layers}
\begin{tabular}{ c | c || c c c c c c c c c c}
\toprule[1pt]
\multirow{2}{*}{Data}     &  \multirow{2}{*}{Metrics} &  \multicolumn{9}{c}{Number of Model Layers}  \\
\cline{3-11}
     &   &  { 1}&  { 2}&  { 3} &  {4} & {5} & {6} & {7} & {8} & {9}  \\
\hline\hline
   &  OA(\%)       &61.79  & 62.28  &74.66 &82.53 &83.84 &84.02 &{\bf 86.25} &84.97 &85.93  \\
  IP &  AA(\%)     &67.42  & 71.30  &84.30 &88.33 &90.58 &90.24 &91.36 &90.25 &{\bf 91.59}  \\
   &  $\kappa$(\%) &56.56  & 57.44  &71.36 &80.03 &81.61 &81.80 &{\bf 84.30} &82.88 &83.97  \\
\hline
   &  OA(\%)       &  84.49  &83.64 &83.89 &85.49 &{\bf 88.38} &87.70&87.19 &86.91 & 86.90  \\
  HS &  AA(\%)     &  84.15  &84.95 &83.87 &85.71 &88.21 &{\bf 88.70}&88.07 &87.65 & 87.17  \\
   &  $\kappa$(\%) &  83.17  &82.24 &82.50 &84.24 &{\bf 87.38} &86.65&86.08 &85.78 & 85.76  \\
\bottomrule[1pt]
\end{tabular}
\end{center}
\end{table*}

\subsection{Performance with Different Spectral Neighbors}
In addition to the number of model layers, different spectral neighbors to construct the spectral tokens in the spectral transformer also play an important role in the classification performance.
To thoroughly evaluate the performance of proposed MultiFormer, in this set of experiments, we will investigate the parameter sensitivity over both the datasets. 
We choose the number spectral neighbors from the set of $\{1,3,5,7,9,11\}$.
Table \ref{table:neighbors} quantifies the classification performance comparison of using different number of spectral neighbors on both the Indian Pines and Houston2013 data.

From the table, we can find that a proper spectral neighbors can guarantee a good performance of the proposed MultiFormer. 
In general, more spectral neighbors would improve the inforation utilization efficiency and could improve the performance for most of the datasets.
While excessive number of spectral neighbors multiplies the computational complexity and further increases the difficulty in model training.
For Indian Pines data, the proposed method performs the best when the number of spectral neighbors is set to 1. Under more spectral neighbors, the performance significantly decreases. 
As for Houston2013 data, the proposed method can achieve 88.38\% when the number of spectral neighbors is set to 5.

\begin{table}[t]
\begin{center}
\caption{Classification accuracies (OA, AA, and $\kappa$) of the proposed method with different number of spectral neighbors.}
\label{table:neighbors}
\begin{tabular}{ c | c || c c c c c c c}
\toprule[1pt]
\multirow{2}{*}{Data}     &  \multirow{2}{*}{Metrics} &  \multicolumn{6}{c}{Number of spectral neighbors}  \\
\cline{3-8}
     &   &  { 1}&  { 3}&  { 5} &  { 7} & { 9} & { 11}  \\
\hline\hline
   &  OA(\%)        &  {\bf 85.96} &83.78 &83.84 &82.53 &82.06 &82.51   \\
  IP &  AA(\%)      &  89.68 &89.69 &{\bf 90.58} &89.38 &88.41 &89.51   \\
   &  $\kappa$(\%)  &  {\bf 83.86} &81.51 &81.61 &80.15 &79.54 &80.11   \\
\hline
   &  OA(\%)       & 85.16  & 85.50&{\bf 88.38} &85.79 &85.77 &84.80    \\
  HS &  AA(\%)     & 85.35  & 85.90& {\bf 88.21} &86.92 &86.19 &84.40    \\
   &  $\kappa$(\%) & 83.88  & 84.25&{\bf 87.38} &84.58 &84.54 &83.49    \\
\bottomrule[1pt]
\end{tabular}
\end{center}
\end{table}

\subsection{Performance with Different Spectral Embedding Dimension $d_2$}
To further validate the classification performance of the proposed MultiFormer with different spectral embedding dimensions which denote the output dimension of the multiscale spatial convolutional transformer, this subsection would conduct additional experiments with different patch dimensions over both the datasets. The number of spectral embedding dimensions is selected from $\{16, 32, 64, 128, 256, 512\}$. The average results with the standard deviation values in terms of the OA obtained by the proposed MultiFormer are reported in Table \ref{table:patches}. There is a basically reasonable trend in OA's results (see table \ref{table:patches}).
That is, with the increase of the number of spectral embedding dimensions, the classification performance gradually improves. While under excessive large number of spectral embedding dimensions, the classification performance would result in the decrease of the classification performance. As shows in the table, for Indian Pavia data, the proposed method can obtain an accuracy of 85.91\% OA with the patch dimension of 512, which performs the best. While for Houston2013 data, the proposed method provide the best performance (88.38\% OA) when the patch dimension is set to 64. The reason is that larger spectral embedding dimension increases the representational ability of the model while excessive large spectral embedding dimension increases the computational complexity which in turn decrease the classification performance. 

\begin{table}[t]
\begin{center}
\caption{Classification accuracies (OA, AA, and $\kappa$) of the proposed method with different spectral embedding dimensions.}
\label{table:patches}
\begin{tabular}{ c | c || c c c c c c c}
\toprule[1pt]
\multirow{2}{*}{Data}     &  \multirow{2}{*}{Metrics} &  \multicolumn{6}{c}{Spectral Embedding Dimension}  \\
\cline{3-8}
     &   &  { 16}&  {32}&  {64} &  {128} & {256} & {512}  \\
\hline\hline
   &  OA(\%)        &  {81.01} &83.45 &83.84 &83.53 &83.73 &{\bf 85.91}   \\
  IP &  AA(\%)      &  88.31 &89.84 &{90.58} &90.29 &90.70 &{\bf 91.73}   \\
   &  $\kappa$(\%)  &  {78.40} &76.16 &81.61 &81.25 &81.43 &{\bf 83.95}   \\
\hline
   &  OA(\%)       &  85.76 &86.55 &{\bf 88.38} &86.90 &85.03 & 84.41   \\
  HS &  AA(\%)     &  86.83 &87.32 &{\bf 87.37} &87.37 &86.10 & 85.62   \\
   &  $\kappa$(\%) &  84.54 &85.40 &{\bf 85.78} &85.78 &83.75 & 83.08   \\
\bottomrule[1pt]
\end{tabular}
\end{center}
\end{table}

\subsection{Comparison with State-of-the-art Methods}

Quantitative classification results in terms of three indices, i.e., OA, AA, and $\kappa$, and the accuracies per class are listed in tables \ref{table:indian_comparison} and \ref{table:houston2013_comparison} for Indian Pines data, and Hosuton2013 data, respectively.
Overall, the conventional classifier, e.g., SVM, and classic backbone networks, such as CNNs-based models (3-D CNN, PResNet, HybridSN), RNNs-based models(RNN), GCNs-based models(miniGCN), and Transformers-based models(ViT, SpectralFormer, SSFTTNet), are chosen as baselines to validate the effectiveness of the proposed method. 

From table \ref{table:indian_comparison}, we can obtain that the proposed method can obtain an accuracy of 87.52\% which performs better than that by SVM (76.53\%), CNNs(e.g., 3-D CNN(77.22\%),  PResNet (82.97\%), HybridSN (78.72\%)), RNN (81.11\%), miniGCN (74.71\%) as well as other state-of-the-art transformers (e.g., ViT (65.16\%), SpectralFormer (83.38\%), SSFTTNet (80.29\%)).
Besides, as shows in table \ref{table:houston2013_comparison}, the proposed method can obtain an accuracy of 88.38\% outperforms the SVM (80.16\%), 3-D CNN (84.71\%), PResNet (85.59\%), HybridSN (86.89\%), RNN (83.55\%), miniGCN (82.31\%), ViT (82.22\%), SpectralFormer(85.55\%), and SSFTTNet (82.46\%).  These comparison results show the effectiveness of the proposed method.

\begin{figure*}[t]
\centering
 \subfigure[]{\label{subfig:pavia}\includegraphics[width=0.16\linewidth]{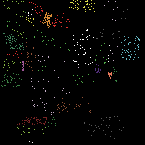}}
 \subfigure[]{\label{subfig:pavia_label}\includegraphics[width=0.16\linewidth]{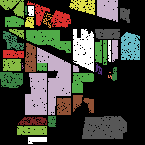}}
 \subfigure[]{\label{subfig:mapping}\includegraphics[width=0.16\linewidth]{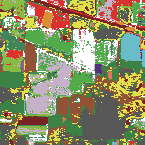}}
\subfigure[]{\label{subfig:mapping}\includegraphics[width=0.16\linewidth]{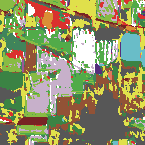}}
\subfigure[]{\label{subfig:mapping}\includegraphics[width=0.16\linewidth]{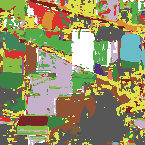}}
\subfigure[]{\label{subfig:mapping}\includegraphics[width=0.16\linewidth]{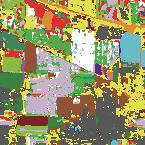}}
\subfigure[]{\label{subfig:mapping}\includegraphics[width=0.16\linewidth]{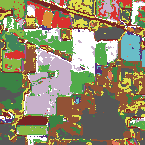}}
\subfigure[]{\label{subfig:mapping}\includegraphics[width=0.16\linewidth]{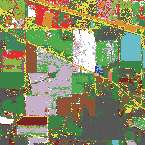}}
\subfigure[]{\label{subfig:mapping}\includegraphics[width=0.16\linewidth]{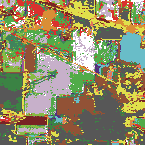}}
\subfigure[]{\label{subfig:mapping}\includegraphics[width=0.16\linewidth]{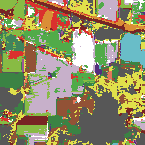}}
\subfigure[]{\label{subfig:mapping}\includegraphics[width=0.16\linewidth]{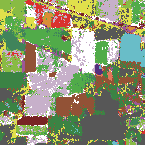}}
\subfigure[]{\label{subfig:mapping}\includegraphics[width=0.16\linewidth]{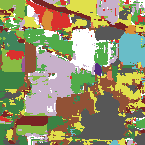}}

   \caption{Indian Pines data. (a) Training; (b) Testing; (c) SVM; (d) 3-D CNN; (e) PResNet; (f) HybridSN; (g) RNN; (h) miniGCN; (i) ViT; (j) SpectralFormer; (k) SSFTTNet; (l) MultiscaleFormer.}
\label{fig:indian_comparison}
\end{figure*}

\begin{figure*}[t]
\centering
 \subfigure[]{\label{subfig:pavia}\includegraphics[width=0.49\linewidth]{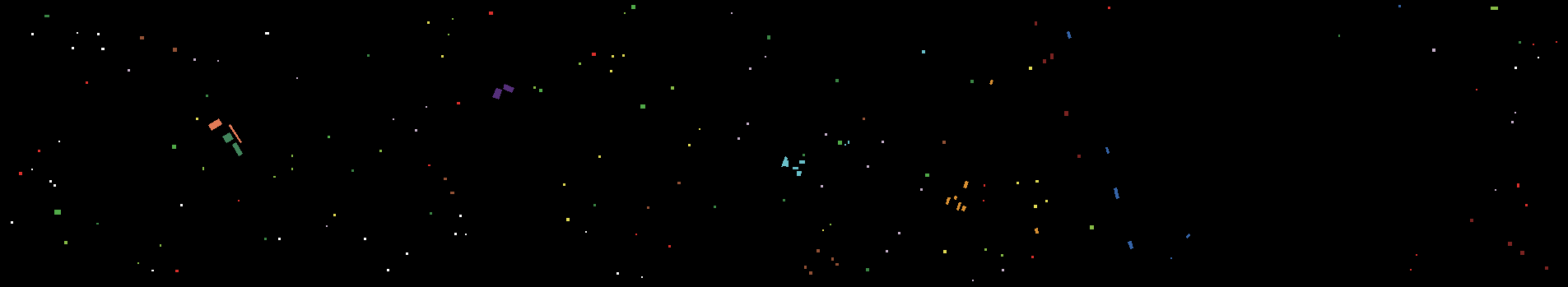}}
 \subfigure[]{\label{subfig:pavia_label}\includegraphics[width=0.49\linewidth]{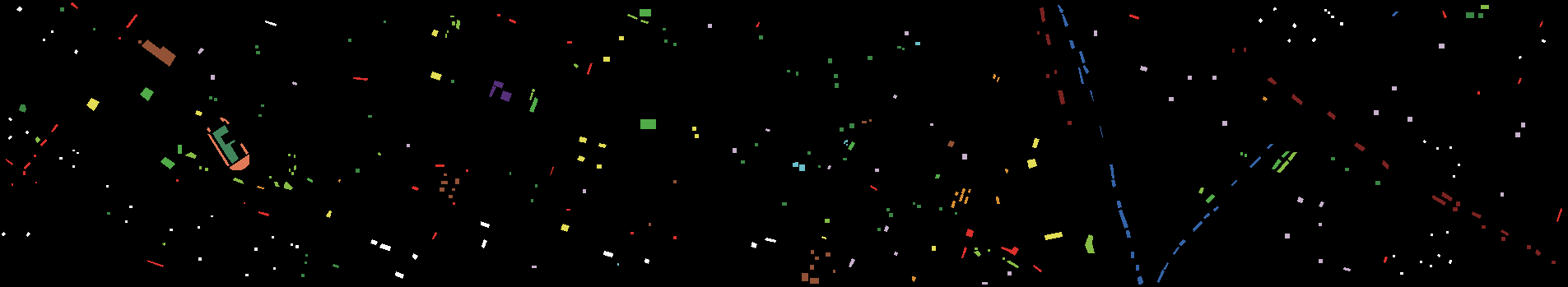}}
 \subfigure[]{\label{subfig:mapping}\includegraphics[width=0.49\linewidth]{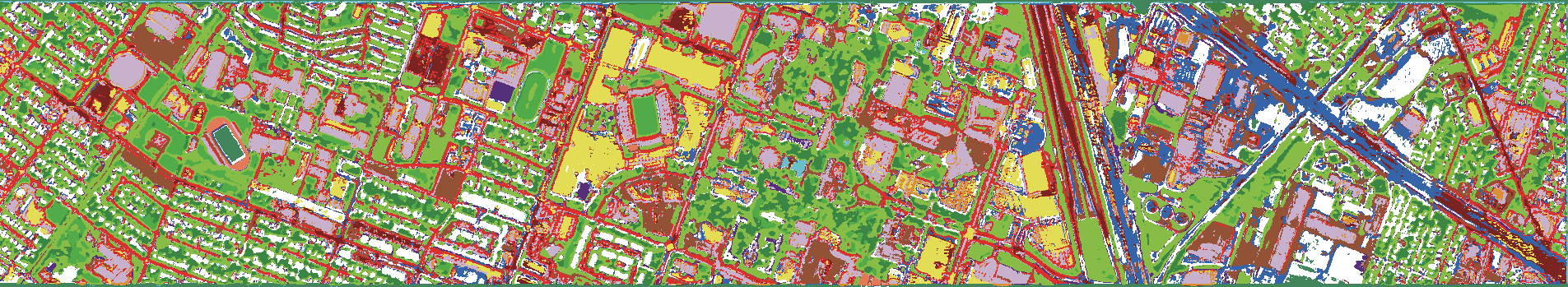}}
\subfigure[]{\label{subfig:mapping}\includegraphics[width=0.49\linewidth]{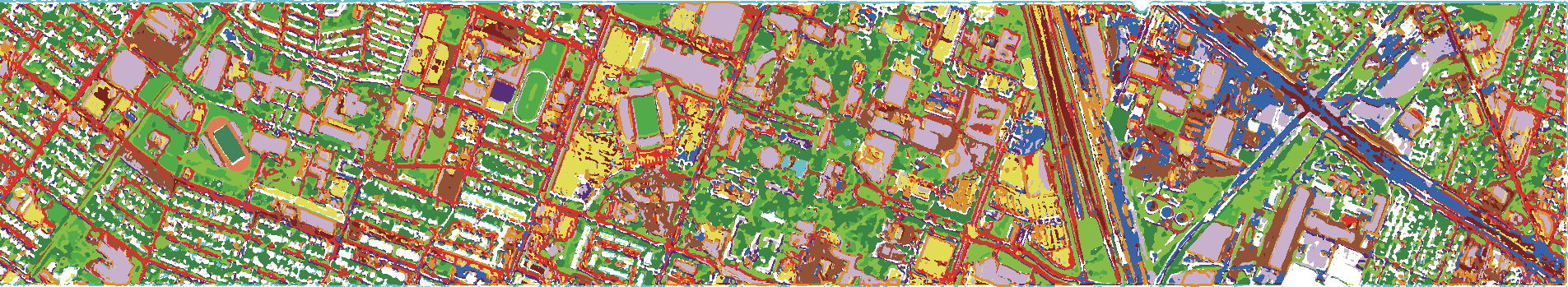}}
\subfigure[]{\label{subfig:mapping}\includegraphics[width=0.49\linewidth]{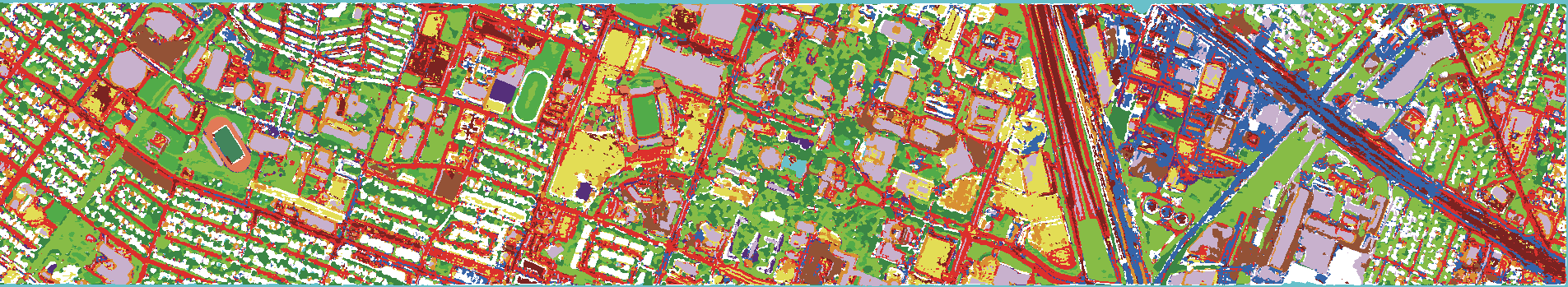}}
\subfigure[]{\label{subfig:mapping}\includegraphics[width=0.49\linewidth]{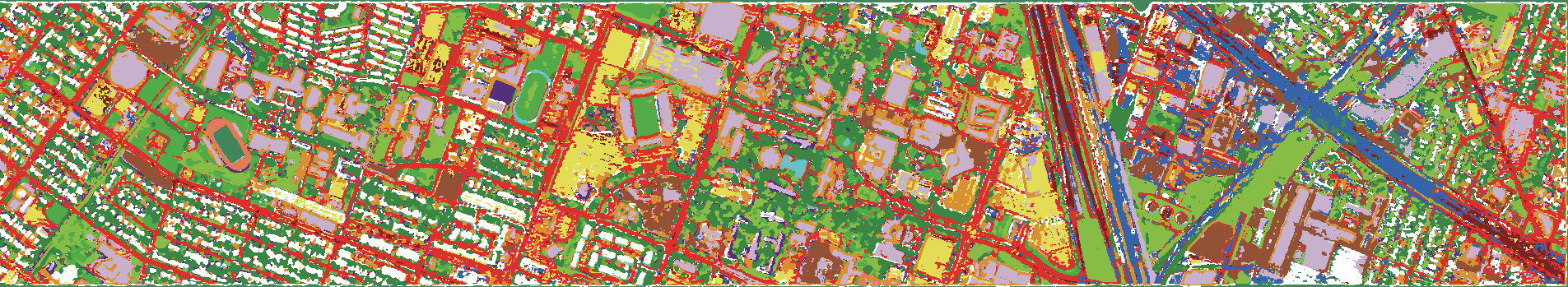}}
\subfigure[]{\label{subfig:mapping}\includegraphics[width=0.49\linewidth]{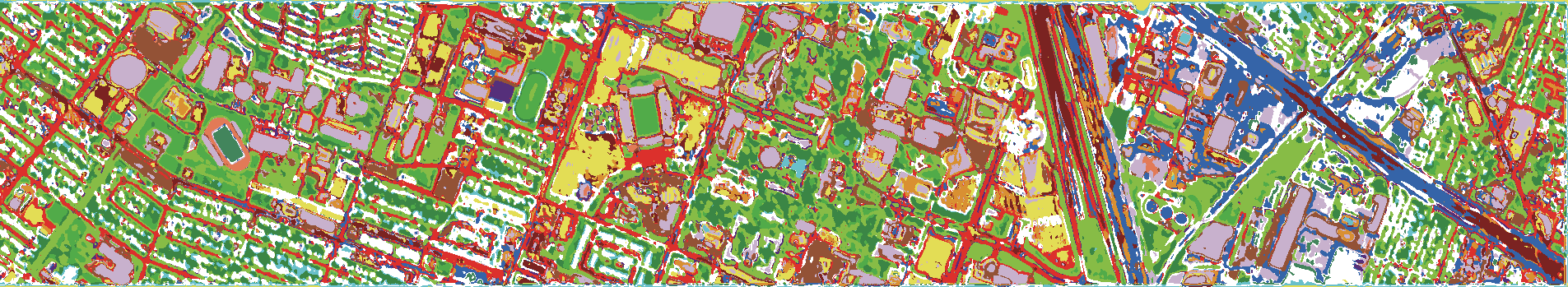}}
\subfigure[]{\label{subfig:mapping}\includegraphics[width=0.49\linewidth]{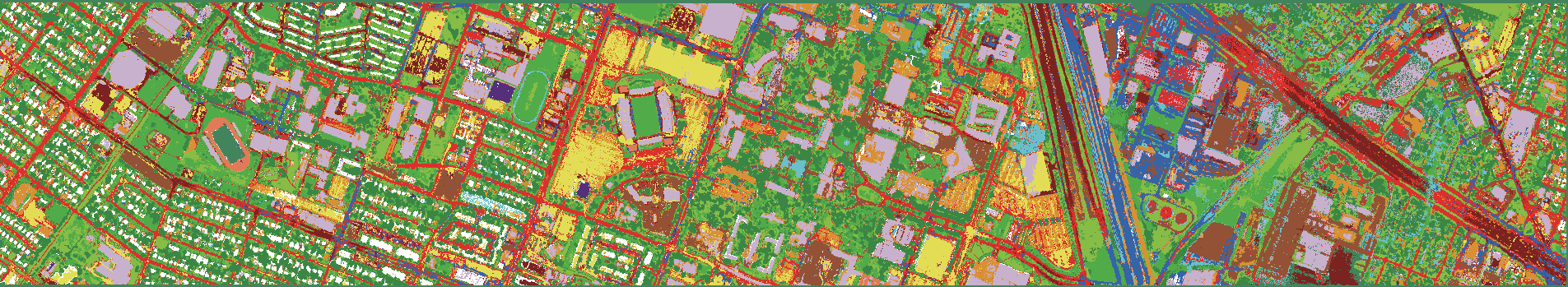}}
\subfigure[]{\label{subfig:mapping}\includegraphics[width=0.49\linewidth]{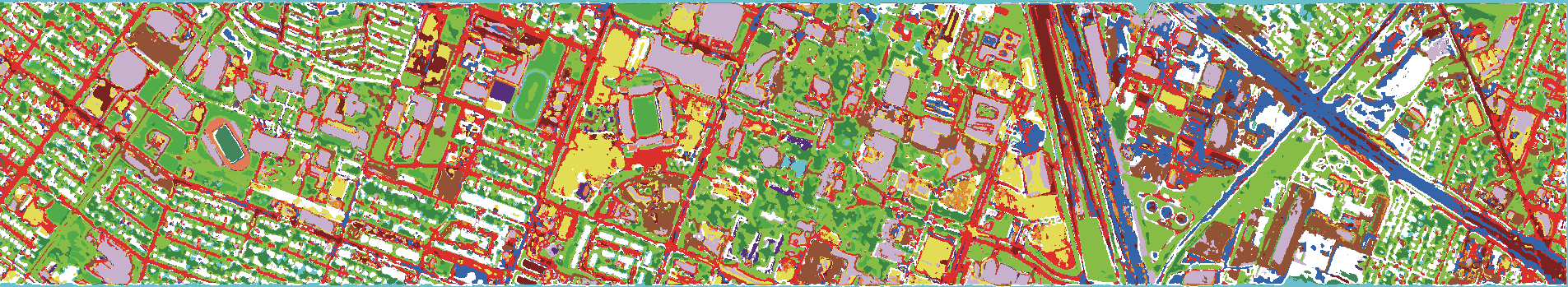}}
\subfigure[]{\label{subfig:mapping}\includegraphics[width=0.49\linewidth]{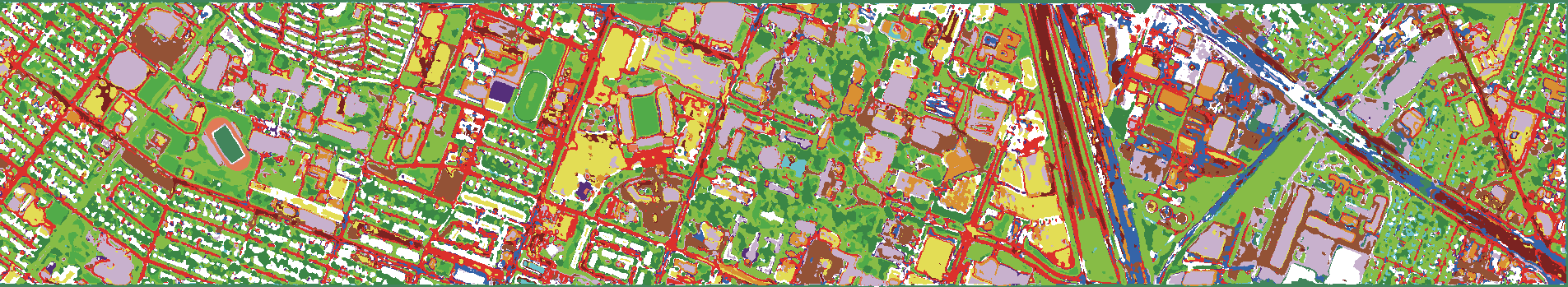}}
\subfigure[]{\label{subfig:mapping}\includegraphics[width=0.49\linewidth]{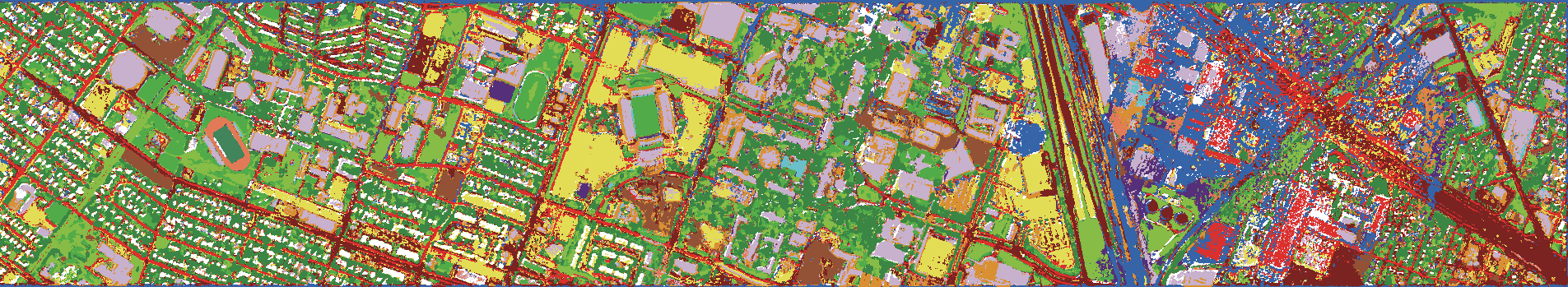}}
\subfigure[]{\label{subfig:mapping}\includegraphics[width=0.49\linewidth]{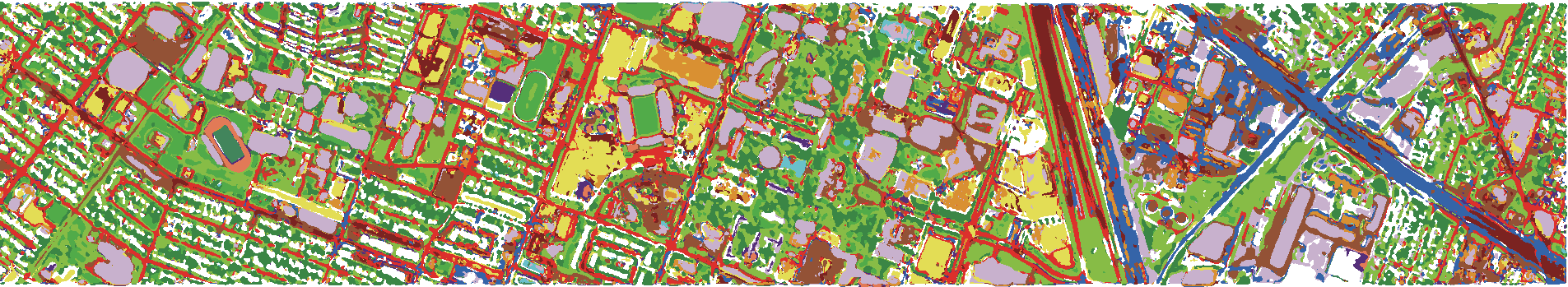}}

   \caption{Houston2013 data. (a) Training; (b) Testing; (c) SVM; (d) 3-D CNN; (e) PResNet; (f) HybridSN; (g) RNN; (h) miniGCN; (i) ViT; (j) SpectralFormer; (k) SSFTTNet; (l) MultiscaleFormer.}
\label{fig:houston2013_comparison}
\end{figure*}

Besides, qualitative evaluation by visualizing the classification maps of different methods is provided over the two datasets. Figs. \ref{fig:indian_comparison} and \ref{fig:houston2013_comparison} shows the visualization results over Indian Pines data and Houston2013 data, respectively.  It is clear from these comparison results that the proposed method can significantly decrease the classification error. In particlular, the results of our methods have less noisy points compared with other state-of-the-art methods.

Overall, the proposed method which utilizes the multiscale spatial-spectral information can provide a better classification performance than other state-of-the-art methods for hyeprspectral image.
%

\begin{table*}[t]
\setlength{\aboverulesep}{-0.1pt}
\setlength{\belowrulesep}{-0.1pt}
\begin{center}
\caption{Classification accuracies (OA, AA, and $\kappa$) of different methods achieved on the Indian Pines data. (spectral neighbors=1, depth=7)}
\label{table:indian_comparison}
\begin{tabular}{ c || c | c c c | c | c | c c c || c}
\toprule[1pt]
\multirow{2}{*}{Methods}     &  \multirow{2}{*}{SVM} &  \multicolumn{3}{c|}{CNNs} &  \multirow{2}{*}{RNN} & \multirow{2}{*}{miniGCN} & \multicolumn{3}{c||}{Transformers} &\multirow{2}{*}{ MultiscaleFormer} \\
 \cmidrule(lr){3-5} \cmidrule(lr){8-10}
    &   &  { 3-D CNN}&  {PResNet }&  { HybridSN} &   &  & { ViT} & {SpectralFormer} & SSFTTNet & \\
\hline\hline
 C1   & 72.11 &67.85     &72.76 & 66.04&72.76 &70.52  &59.68 &78.97 &{\bf 83.37} &74.13 \\
 C2   & 71.43 &77.04     &87.50 & 79.46&84.82 &53.19  &37.76 &85.08 &71.39 &{\bf 89.67} \\
 C3   & 86.96 &93.48     &94.02 & {\bf 94.57}&78.26 &91.85  &55.43 &85.33 &94.57 &92.93 \\
 C4   & 95.97 &92.84     &94.85 & 91.28&88.14 &93.74  &65.10 &94.18 &93.93 &{\bf 97.09} \\
 C5   & 88.67 &83.21     &91.97 & 90.96&83.50 &95.12  &86.51 &84.36 &{\bf 95.38} &95.27 \\
 C6   & 95.90 &98.63     &97.49 & {\bf100.0}&91.80 &99.09  &97.95 &98.63 &98.40 &98.86 \\
 C7   & 75.60 &74.51     &84.20 & 79.30&87.58 &63.94  &51.85 &63.94 &72.66 &{\bf 84.97} \\
 C8   & 59.02 &62.66     &73.57 & 68.07&74.28 &68.40  &62.45 &83.95 &71.63 &{\bf 84.37} \\
 C9   & 76.77 &69.68     &75.18 & 70.21&75.35 &73.40  &41.13 &73.58 &50.18 &{\bf 79.08} \\
 C10   &99.38 &99.38     &100.0 & {\bf 100.0}&99.38 &98.77  &96.30 &99.38 &98.15 &99.38 \\
 C11   &93.33 &93.25      &93.41 & 76.94&93.89 &88.83  &91.00 &97.19 &91.94 &{\bf 95.26} \\
 C12   &73.94 &96.06      &80.61 & 90.30&63.03 &46.06  &52.12 &64.55 &86.63 &{\bf 97.88} \\
 C13   &100.0 &100.0      &100.0 & 100.0&100.0 &97.78  &95.56 &97.78 &95.45 &{\bf 100.0} \\
 C14   &87.18 &89.74      &{\bf 97.44} & 92.31&66.67 &46.15  &48.72 &76.92 &82.05 &92.31 \\
 C15   &100.0 &90.91      &100.0 & 100.0&100.0 &72.73  &81.82 &100.0 &100.0 &{\bf 100.0} \\
 C16   &100.0 &100.0      &100.0 & 100.0&100.0 &80.00  &100.0 &100.0 &100.0 &{\bf 100.0} \\
                                                                                                             
 \hline\hline
 {OA(\%)}   & 76.53   &  77.22   &82.97  & 78.72&81.11 &74.71 &65.16 &83.38  &80.29  &{\bf 87.52} \\
 
 {AA(\%)}   & 86.02   &  86.83   &90.19  & 88.15&84.97 &77.47 &70.21 &86.49  &86.61  &{\bf 92.57} \\
 
 {$\kappa$(\%)}&73.42 & 74.21    &80.65  & 75.81&78.51 &71.21 &60.26 &80.93  &77.40  &{\bf 85.74} \\
\bottomrule[1pt]
\end{tabular}
\end{center}
\end{table*}

\begin{table*}[t]
\setlength{\aboverulesep}{-0.1pt}
\setlength{\belowrulesep}{-0.1pt}
\begin{center}
\caption{Classification accuracies (OA, AA, and $\kappa$) of different methods achieved on the Houston 2013 data.}
\label{table:houston2013_comparison}
\begin{tabular}{ c || c | c c c | c | c | c c c || c}
\toprule[1pt]
\multirow{2}{*}{Methods}     &  \multirow{2}{*}{SVM} &  \multicolumn{3}{c|}{CNNs} &  \multirow{2}{*}{RNN} & \multirow{2}{*}{miniGCN} & \multicolumn{3}{c||}{Transformers} &\multirow{2}{*}{ MultiscaleFormer} \\
 \cmidrule(lr){3-5} \cmidrule(lr){8-10}
    &   &  { 3-D CNN}&  {PResNet }&  { HybridSN} &   &  & { ViT} & {SpectralFormer} & SSFTTNet & \\
\hline\hline
 C1   &  82.62  & 83.76  & 81.67&83.57 & 81.67&{\bf 96.20}  &82.53 &83.29 & 83.29& 87.27\\
 C2   &  98.78  & 95.49  & 99.91&{\bf 100.0} & 95.39&96.90  &99.06 &98.97 &90.51 & 96.15\\
 C3   &  90.30  & 95.05  & 90.89&98.02 & 95.05&{\bf 99.41}  &91.49 &96.63 &98.61 & 85.94\\
 C4   &  97.06  & {\bf 99.24}  & 86.74&95.55 & 96.02&97.63  &95.64 &96.02 &96.97 & 98.11\\
 C5   &  99.81  & 99.43  & 99.43&99.72 & 97.63&97.73  &99.34 &{\bf 100.0} &99.53 & 99.91\\
 C6   &  82.52  & 90.21  & 92.31&{\bf 95.80} & 91.61&95.10  &94.41 &94.40 &91.61 & 93.71\\
 C7   &  89.65  & 86.85  & 90.49&90.67 & 89.92&65.86  &{\bf 91.04} &83.21 &67.91 & 89.37\\
 C8   &  57.74  & 82.05  & 75.31&81.01 & 70.09&65.15  &60.68 &80.72 &55.08 & {\bf 82.62}\\
 C9   &  61.19  & 76.49  & {\bf 80.93}&81.59 & 73.84&69.88  &71.20 &77.43 &54.25 & 80.08\\
 C10   & 67.66   & 53.96  &70.27 &46.33 &65.93 &67.66  &52.51 &58.01 &{\bf 81.56} & 70.08\\
 C11   & 72.68	   & 82.35  &84.91 &{\bf 94.12} &70.40 &82.83  &78.75 &80.27 &90.51 & 88.90\\
 C12   & 70.41   & 78.48  &71.85 &80.50 &79.73 & 68.40 &81.27 &84.44 &84.73 & {\bf 91.16}\\
 C13   & 61.05   & 75.44  &89.47 &{\bf 94.74} &74.39 & 57.54 &65.96 &73.33 &81.75 & 63.16\\
 C14   & 94.33   & 91.90  &97.57 &96.36 &98.79 & 99.19 &95.14 &{\bf 99.60} &99.19 & 97.16\\
 C15   & 80.13   & 92.18  &{\bf 100.0} &95.78 &98.31 & 98.73 &92.39 &99.15 &99.79 & 99.58\\
                                                                                                             
 \hline\hline
 {OA(\%)}      & 80.16  & 84.71 & 85.59 &86.89 &83.55 &82.31 &82.22 & 85.55 & 82.46 & {\bf 88.38}\\
 
 {AA(\%)}      & 80.40  & 85.53 & 87.45 &{\bf 88.92} &85.25 &83.88 &83.43 & 87.03 & 85.02 & 88.21\\
 
 {$\kappa$(\%)} & 78.44   & 83.40 & 84.35 &85.77 &82.15 &80.84 &80.68 & 84.32 & 80.97 & {\bf 87.38}\\
\bottomrule[1pt]
\end{tabular}
\end{center}
\end{table*}

\section{Conclusions}
In this paper, a multiscale spectral-spatial convolutional Transformer, which embeds a multiscale spatial convolutional transformer in spectral transformer, is developed for hyperspectral image classification. First, this work develops the multiscale spatial convolutional transformer as inner transformer block to extract the multiscale spatial information. Then, the spectral transformer, which is designed to extract the long-range spectral information, act as the outer transformer block. By stacking the inner and outer blocks, and using spectral-spatial cross-layer adaptive fusion module to fuse the cross-layer spatial and spectral information, the MultiscaleFormer can be constructed to extract discriminative multiscale spectral-spatial information. Experimental results show the advantage of the proposed architecture when compared with other state-of-the-art methods.

In the future, we will investigate strategies to reduce the complexity of Transformer structures and make the models more efficient, accessible, and applicable in real world application. Moreover, we would also like to explore novel training methods tailored to the unique characteristics of the proposed MultiFormer structure, with a particular focus on embedding hyperspectral image intrinsic features.

\ifCLASSOPTIONcaptionsoff
  \newpage
\fi



%

\bibliographystyle{IEEEtran}
\bibliography{references}




%




%


\vfill


\end{document}